\documentclass[sigconf]{aamas} 

\usepackage{balance} 
\usepackage{amsthm}

\setcopyright{ifaamas}
\acmConference[AAMAS '23]{Proc.\@ of the 22nd International Conference
on Autonomous Agents and Multiagent Systems (AAMAS 2023)}{May 29 -- June 2, 2023}
{London, United Kingdom}{A.~Ricci, W.~Yeoh, N.~Agmon, B.~An (eds.)}
\copyrightyear{2023}
\acmYear{2023}
\acmDOI{}
\acmPrice{}
\acmISBN{}

\acmSubmissionID{1103}

\title[AAMAS-2023 Traffic Signal Control]{SocialLight: Distributed Cooperation Learning towards Network-Wide Traffic Signal Control}

\author{Harsh Goel}
\affiliation{
  \institution{University of Pennsylvania}
  \city{Philadelphia}
  \country{United States}}
\email{harshg99@seas.upenn.edu}

\author{Yifeng Zhang}
\affiliation{
  \institution{National University of Singapore}
  \city{Singapore}
  \country{Singapore}}
\email{yifeng@u.nus.edu}

\author{Mehul Damani}
\affiliation{
  \institution{Massachusetts Institute of Technology}
  \city{Cambridge}
  \country{United States}}
\email{damanimehul24@gmail.com}

\author{Guillaume Sartoretti}
\affiliation{
  \institution{National University of Singapore}
  \city{Singapore}
  \country{Singapore}}
\email{guillaume.sartoretti@nus.edu.sg}

\begin{abstract}

Many recent works have turned to multi-agent reinforcement learning (MARL) for adaptive traffic signal control to optimize the travel time of vehicles over large urban networks. However, achieving effective and scalable cooperation among junctions (agents) remains an open challenge, as existing methods often rely on extensive, non-generalizable reward shaping or on non-scalable centralized learning. To address these problems, we propose a new MARL method for traffic signal control, SocialLight, which learns cooperative traffic control policies by distributedly estimating the individual marginal contribution of agents on their local neighborhood. SocialLight relies on the Asynchronous Actor Critic (A3C) framework, and makes learning scalable by learning a locally-centralized critic conditioned over the states and actions of neighboring agents, used by agents to estimate individual contributions by counterfactual reasoning. We further introduce important modifications to the advantage calculation that help stabilize policy updates. These modifications decouple the impact of the neighbors' actions on the computed advantages, thereby reducing the variance in the gradient updates. We benchmark our trained network against state-of-the-art traffic signal control methods on standard benchmarks in two traffic simulators, SUMO and CityFlow. Our results show that SocialLight exhibits improved scalability to larger road networks and better performance across usual traffic metrics.

\end{abstract}


\newcommand{\BibTeX}{\rm B\kern-.05em{\sc i\kern-.025em b}\kern-.08em\TeX}

\usepackage{caption}
\usepackage[caption=false]{subfig}

\begin{document}


\pagestyle{fancy}
\fancyhead{}


\maketitle


\section{Introduction}
\label{introduction}
In recent years, most cities around the world have seen growing traffic levels, and associated traffic congestion have started to show a number of negative effects, both at the micro and macro level.
At the micro level, passengers experience frustration due to delays and also face increased risks of collisions.
At the macro level, unproductive time spent in traffic damages economic health, while wasted fuel and traffic jams increase air and noise pollution.
As a result, there is a growing need for effective traffic signal control methods, which can play a significant role in alleviating traffic congestion.

Current traffic signal control methods can be broadly classified into two categories: fixed-time control and adaptive control.
In fixed-time control, the duration of different traffic light phases are pre-determined, often optimized offline from historical data.
However, urban traffic, on top of having considerable stochasticity, also shows significant temporal and spatial variations.
For example, higher congestion is often seen due to temporal peaks at the end of a work day.
Similarly, the spatial structure of traffic networks often gives rise to tidal patterns, with high congestion in particular lane directions.
To account for this variability, adaptive traffic signal control (ATSC) methods aims at dynamically adjusting traffic signal phases online, based on current traffic conditions.

Multi-Agent Reinforcement learning (MARL) is one such adaptive and versatile data-driven method, which has recently shown great promise in ATSC and general control tasks~\cite{lowe2017multi,chu2019multi,foerster2018counterfactual}.
ATSC is cast as a MARL problem in which each agent controls a single traffic intersection, based on locally-sensed real-time traffic conditions and communication with neighboring intersections.
Thus, each agent learns a policy which maps the current traffic conditions at the intersection into control outputs (e.g., phase selection, phase duration).
This lends it an advantage over conventional ATSC methods, which rely on complex dynamics models and heuristic assumptions.
An alternative to MARL is to train a single centralized RL agent, which is responsible for controlling all traffic intersections.  
However, while centralization allows for direct maximization of a global reward/objective such as average trip time, training such a centralized method is infeasible in practice due to the exponentially growing joint action space, and the high latency associated with information centralization. 

Although the MARL formulation of ATSC alleviates most issues associated with centralized methods, it introduces new challenges as the performance of control policies that optimize local objectives for each agent (intersection) will not be equivalent to that of a centralized global RL agent if the local objectives aren't well-aligned with the global (team-/network-level) one.
Since, traffic networks have complex spatio-temporal patterns and significant interdependence between agents, greedily optimizing each agent's local reward usually does not optimize global (network-level) objectives.
A possible solution to this is to directly sum each agent's local reward with that of neighboring agents into a large \textit{neighborhood reward}, which becomes a new, more global objective optimized by each agent.
The idea here is to couple neighboring agents via their rewards, whereby improving their neighbors' local rewards via their own actions directly affect their own long-term return.
However, such a neighborhood reward has high variance since it is now conditioned on the actions of multiple agents, making it difficult for an agent to determine its true \textit{marginal} contribution.
A recent work introduced COMA~\cite{foerster2018counterfactual}, which learns a complex team-level network allowing agents to estimate their own marginal contribution to the team reward via counterfactual reasoning.
Specifically, COMA's centralized value function network uses both the global team reward, as well as the states and actions of all agents as input, which becomes exponentially harder to train in larger teams.

In this work, we propose to spatially distribute the global credit assignment problem into a collection of local marginal contributions calculations, as a natural means to balance the tradeoff between scalability and cooperative performance.
To this end, we learn a shared value network, similar to COMA's but only conditioned on each agent's states/actions and that of its direct neighbors, which can be used for agents to estimate their own marginal contribution to the local neighborhood reward.
By relying on a fixed number of neighbors, our locally-centralized value network allows for significantly improved scalability, while minimally affecting the quality of the learned solutions by leveraging the natural fixed structure of ATSC, where the natural flow of traffic means that neighboring agents must be more tightly coupled.
We further introduce important modifications to the advantage calculation that helps improve and stabilize policy updates.

We present results of an extensive set of simulations conducted on a range of benchmark traffic networks using two standard traffic simulators, SUMO~\cite{krajzewicz2010traffic} and CityFlow~\cite{tang2019cityflow}.
We show that our framework - SocialLight - results in improved cooperation and natural scalability to larger networks compared to existing state-of-the-art ATSC baselines.
To the best of our knowledge, we are also the first work to show effective performance on both these standard traffic simulators\footnote{
To help the community standardize bench-marking on both simulators, our open-source code can be found at \url{https://github.com/marmotlab/SocialLight}}.
Finally, through a series of ablation studies, we also show that the modified advantages in combination with the counterfactual baseline derived from COMA help improve the speed and stability of training in comparison to vanilla A3C/COMA.

\vspace{-0.2cm}


\section{Related Work}
\label{related_work}
\subsection{Conventional Traffic Signal Control}

Traffic signal control is a versatile problem with many possible objectives to optimize and different scopes of optimization. 
Conventional methods can be broadly categorized into adaptive or fixed-time control based on the ability of the method to adapt to current traffic conditions. 
They are also categorized based on the scope of their optimization - some methods only consider optimization over a single isolated traffic intersection while others consider a network of traffic intersections (multiple intersections). 
Here, we briefly list out seminal works for each of these categories - 
\begin{itemize}
    \item \textbf{Single intersection, Fixed time:} The Webster method~\cite{koonce2008traffic} obtains a closed-form solution for the optimal cycle length and phase split based on a set of modeling assumptions.  
    \item \textbf{Single intersection, Adaptive:} SCATS~\cite{lowrie1990scats} is a popular adaptive-control method, which has even been deployed in numerous urban cities around the world. It takes in pre-defined signal plans and iteratively selects from these traffic signals according to a defined performance measure.
    \item \textbf{Multiple intersections, Fixed time:} GreenWave~\cite{roess2004traffic} optimizes the timing offsets between different intersections to minimize the number of stops for vehicles traveling along a specific direction. 
    \item \textbf{Multiple intersections, Adaptive:} Max-pressure control~\cite{varaiya2013max} addresses the risk of oversaturation at an intersection by balancing queue lengths between neighboring intersections. 
\end{itemize}
This list is not exhaustive and for more details, we refer the reader to the recent survey by Wei et al.~\cite{wei2019survey}. 
While conventional traffic control methods are currently the standard for real-world deployments, they rely on accurate traffic models.

\vspace{-0.1cm}
\subsection{RL-based Traffic Signal Control}

Model-free RL is particularly suitable for ATSC due to its ability to learn from and find structure in large amounts of raw data.
Early works in RL explored different ATSC problem formulations on simplified traffic environments.
Out of these variants, the most common variant has been learning to select the next traffic light phase using a set of features describing the local traffic conditions.~\cite{wiering2004intelligent,genders2016using,prashanth2011reinforcement}.
In contrast, Li et al.~\cite{li2016traffic}, Aslani et al.~\cite{aslani2017adaptive} and Casas et al.~\cite{casas2017deep} focused on learning policies for selecting the traffic signal timing (also known as the phase duration).
While most methods focused on learning policies from a low-dimensional feature space, Mousavi et al.~\cite{mousavi2017traffic} used a CNN to directly map from image snapshots (obtained using a simulator) to the policy for selecting the next traffic phase.
Similarly, Wei at al.~\cite{wei2018intellilight} used both extracted features and real-world images to learn the policy.
Taking note of the fact that traffic intersections in the real world vary greatly, Oroojlooy et al.~\cite{oroojlooy2020attendlight} proposed AttendLight, which uses two attention networks to learn a universal model applicable to intersections with any number of roads, lanes, phases (possible signals), and traffic flow.

Multiple works described above showed that deep RL agents can effectively control individual intersections. Hence, the focus of more recent works has shifted to developing methods for a network of traffic intersections that more closely resemble real-world traffic systems where different traffic intersections are highly interconnected.
From lessons learned through conventional methods such as Greenwave~\cite{roess2004traffic}, it is evident that coordination between these intersections is necessary to achieve effective performance for the network. 
Current coordination methods for multi-agent traffic signal control can be broadly grouped into two - joint action learners and independent learners.
Joint action learners use a single global agent to control the traffic for all intersections~\cite{prashanth2011reinforcement,xie2020iedqn}.
While joint action learners allow for direct optimization of a global objective, they find it difficult to scale beyond a few intersections. 
On the other hand, independent learners train an individual policy for each intersection, while considering other agents/intersections to be a part of the environment~\cite{wei2019presslight,chu2019multi,nishi2018traffic,wei2019colight,oroojlooy2020attendlight}. 
Wei at al.~\cite{wei2019presslight} proposed PressLight, which extends \textit{max pressure}~\cite{varaiya2013max} to multi-agent RL by rewarding each agent for minimizing the pressure at an intersection. 
Scaling this, Chen et al~\cite{chen2020toward} proposed MPLight, a deep MARL framework which uses parameter sharing to train policies using pressure-based objectives for large-scale networks. 
A fundamental challenge with achieving cooperation with independent learners is partial observability, as individual traffic intersections are unable to observe nearby intersections. 
To address this, Chu et al.~\cite{chu2019multi} proposed MA2C, a MARL algorithm that adds fingerprints of its neighbors to each agent's observation for improved observability, and a spatial discount factor to reduce learning difficulty. 
With a similar motivation, Nishi et al.~\cite{nishi2018traffic} used a graph convolutional neural network (GCNN) to automatically extract traffic features between distant intersections.
As an alternative to direct state augmentations or feature extractions, Wei et al.~\cite{wei2019colight} proposed to learn a communication mechanism between different intersections.
Their framework, referred to as Colight, uses graph attentional networks (GAT's) to facilitate communication between intersections.
More recently, Zhang et al.~\cite{attentionlight} further showed performance improvements by learning phase correlation with an attention mechanism over a queue length based state representation.

Finally, making use of the Centralized Training Decentralized Execution (CTDE) paradigm, some works have focused on learning a centralized critic to guide independent policy learning~\cite{chao2022cooperative,van2016coordinated}.

\begin{figure}
\includegraphics[width=6.8cm,height=4cm]{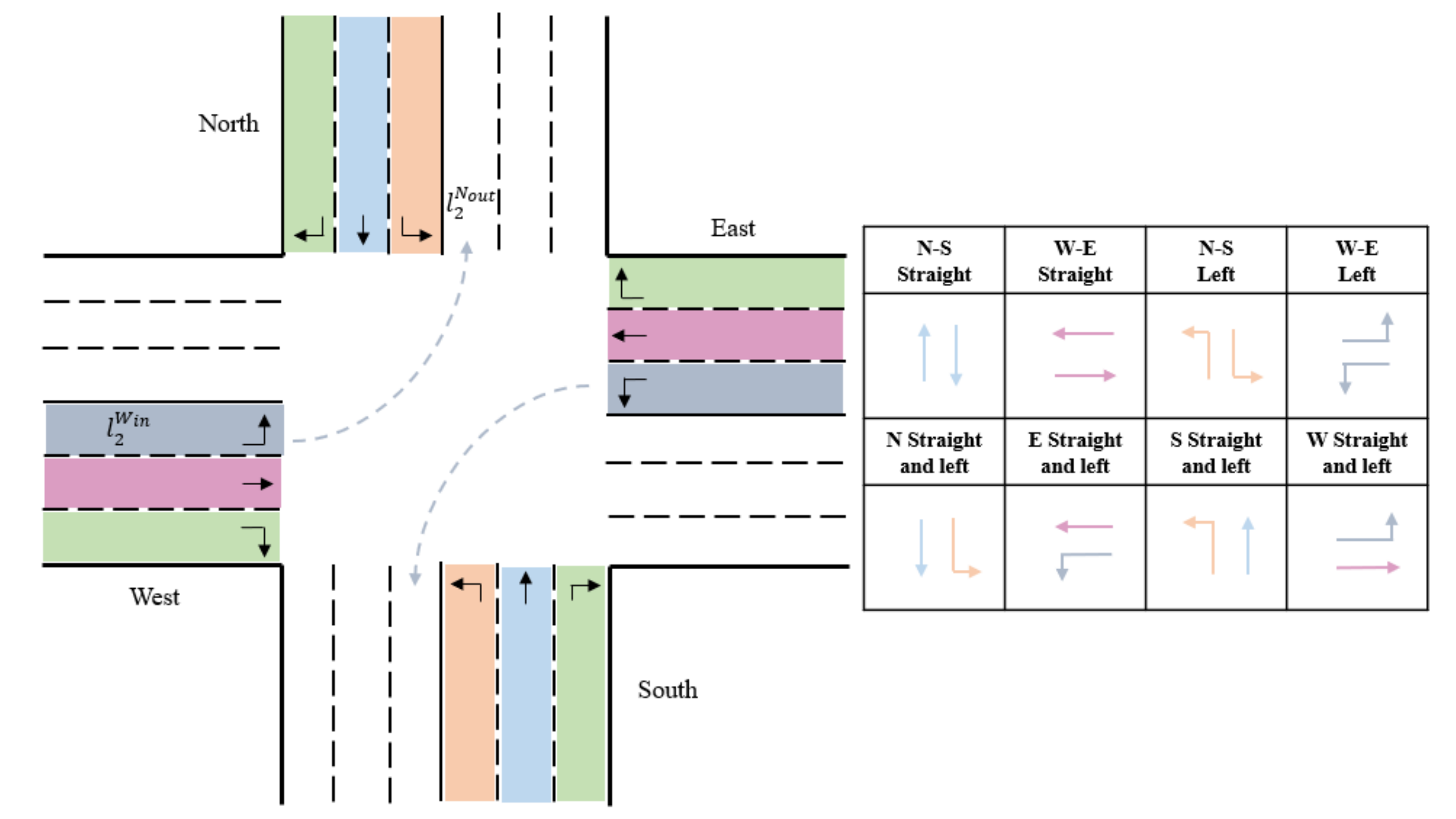}
\caption{Single intersection with 8 traffic light phases} 
\Description{Single intersection with 8 traffic light phases}
\vspace{-0.7cm}
\label{intersection}
\end{figure}

\vspace{-0.1cm}


\section{Background}
\label{premlinaries}
\subsection{Multi-Agent Reinforcement Learning}

Multi-agent Reinforcement Learning generally optimize a global objective over a cooperative game involving numerous agents.
Formally, a MARL problem can be formulated as a set of Decentralized Partially Observable Markov Decision Processes (Dec-POMDPs)~\cite{gupta2017cooperative} characterized by a tuple $G=(\mathbf{\psi}, \mathbf{S}, \mathbf{A}, \mathbf{P}, \mathbf{R}, \mathbf{\rho}, \mathbf{O}, \mathbf{Z}, \mathbf{\gamma})$, where $\mathbf{\psi}$ is a finite set of all agents ($|\psi| = n$, $\mathbf{S}$ is the state space, $A$ the joint action space, defined as $A = A_1 \times A_2 \times .... \times A_n$ and $\mathbf{P} : \mathbf{S} \times \mathbf{A} \times \mathbf{S} \to [0,1]$ denotes the global transition dynamics.
The reward function $R : S\times A \times S \to \mathbb{R}^n$ computes a set of private rewards $ [r^t_i] $ for each agent $i$ at each time-step $t$. These rewards can be global (i.e each agent receives the same global reward) or local via reward shaping.
In partially observable settings each agent cannot access the true global state of the environment $\mathbf{s}_t$. Instead, it draws an observation via observation models $\mathbf{O} = [O_1, O_2, ..., O_N ]$ where $O_i: \mathbf{S}  \rightarrow \mathbf{Z}_i$. Here $ \mathbf{Z} = [Z_1,Z_2,...Z_n]$ are the agents' observation spaces.

Let $\Pi^i : Z_i \times A_i \to [0,1]$ denote the stochastic policy for agent $i$, then the joint policy of the multi-agent system is given by $\pi(\mathbf{a^t}|\mathbf{s^t}) = \prod_{i \in \psi} \pi^i_{\theta} (a^t_i|z^t_i) $ assuming the policy of each agent is parameterized by $\theta_{i}$. The Multi-Agent RL objective thereby is to find an optimal joint policy $\pi$ that formally maximizes the discounted returns over all agents $J(\pi) = \mathop{E}_{\tau \sim \pi}[ \sum_{t=0}^{\infty}\sum _{i=0}^{N} r^t_i]$. Here, $\tau$ denotes the global trajectory $(\mathbf{s}^0,\mathbf{a^0},\mathbf{s}^1,\mathbf{a^1} ..\mathbf{s}^t,\mathbf{a^t})$. Finally, $\mathbf{\rho}$ and $\mathbf{\gamma}$ represents the initial state distribution and the discount factor respectively.

This objective can be optimized over in a centralized manner by parameterizing the global policy $\pi(\mathbf{a^t}|\mathbf{s^t})$.
However, such a centralized approach usually scales poorly, given the exponentially growing state-action space of the agents.
Independent learning algorithms~\cite{mnih2016asynchronous,schulman2017proximal,mappo} have been effective in many multi-agent settings to optimize $\pi^i_{\theta} (a^t_i|z^t_i)$ over the local returns $ {R}^i(\tau) = \sum_{t=0}^{\infty}r^t_i$ with either a local observation value critic $V^{\pi_i}_i(z_{i}) = \mathop{E}_{\tau} [{R}^i(\tau) | z^0_i=z_i]$ or $Q^{\pi_i}_i(z_{i},a_i) = \mathop{E}_{\tau} [{R}^i(\tau)| z^0_i=z_{i},a^0_i=a_i]$ to estimate local advantages $A^{\pi_i}_i(z_{i},a_i) = Q^{\pi_i}_i(z_{i},a_i) - V^{\pi}_i(z_{i})$ for policy improvement. For cooperation, multi-agent actor-critic methods~\cite{maddpg,coma} propose to learn centralized critics $Q^{\pi}(\mathbf{s},\mathbf{a})$ that optimize global returns ${R}(\tau) = \mathop{E}_{\tau}[\sum_{t=0}^{\infty}\sum_{i=0}^{N} r^t_i]$ with individual policies for decentralized execution in their environments.

\vspace{-0.25cm}
\subsection{Traffic Terminology}

\textit{Definition 1 (Traffic movement)}: One way by which vehicles can traverse the intersection, i.e., from one incoming lane to one connected outgoing lane. The traffic movement $m_{ij}$ between lane $i$ and outgoing lane $j$ is denoted as $(l_i^{in}, l_j^{out})$, and the activation of the movement is defined as $m_{ij}=1$.

\textit{Definition 2 (Traffic signal phase)}: A set of simultaneously allowed traffic movements, allowing only vehicles under these activated traffic movements to traverse the intersection. We denote the signal phase as $p=\left\{ m_{ij} | m_{ij}=1 \right\} $, where $i\in \mathbf{L_{in}}$ and $j \in \mathbf{L_{out}}$. 

\textit{Definition 3 (Traffic Agent and traffic network)}: A traffic agent is in charge of one intersection and relies on the real-time traffic conditions within its own area to control the signal phases. A traffic network is a multi-agent network $\mathcal{G}(\mathcal{V},\mathcal{E})$, where the vertices $\mathcal{V}$ are traffic agents and the edges $\mathcal{E}$ define the road network connecting them. An agent $i$ has an immediate neighbor $j$ if $E_{ij}$ = 1. In practice, this means that agents $i$ and $j$ are directly connected.

Fig.~\ref{intersection} depicts an example single intersection which is composed of twelve incoming lanes and twelve outgoing lanes.
Considering a single connection between incoming and outgoing lanes, i.e., each incoming lane is only connected to one outgoing lane, there is a total of twelve movements (left-turn, go-straight, and right-turn for each direction).
Therefore, we can define eight phases, shown in the right side of the Fig.~\ref{intersection}.
Currently, the W-E left-turn phase is activated at the intersection, allowing vehicles at the left-turn lanes of the west and east directions to move.

\begin{figure*}
    \centering
    \includegraphics[width=\linewidth]{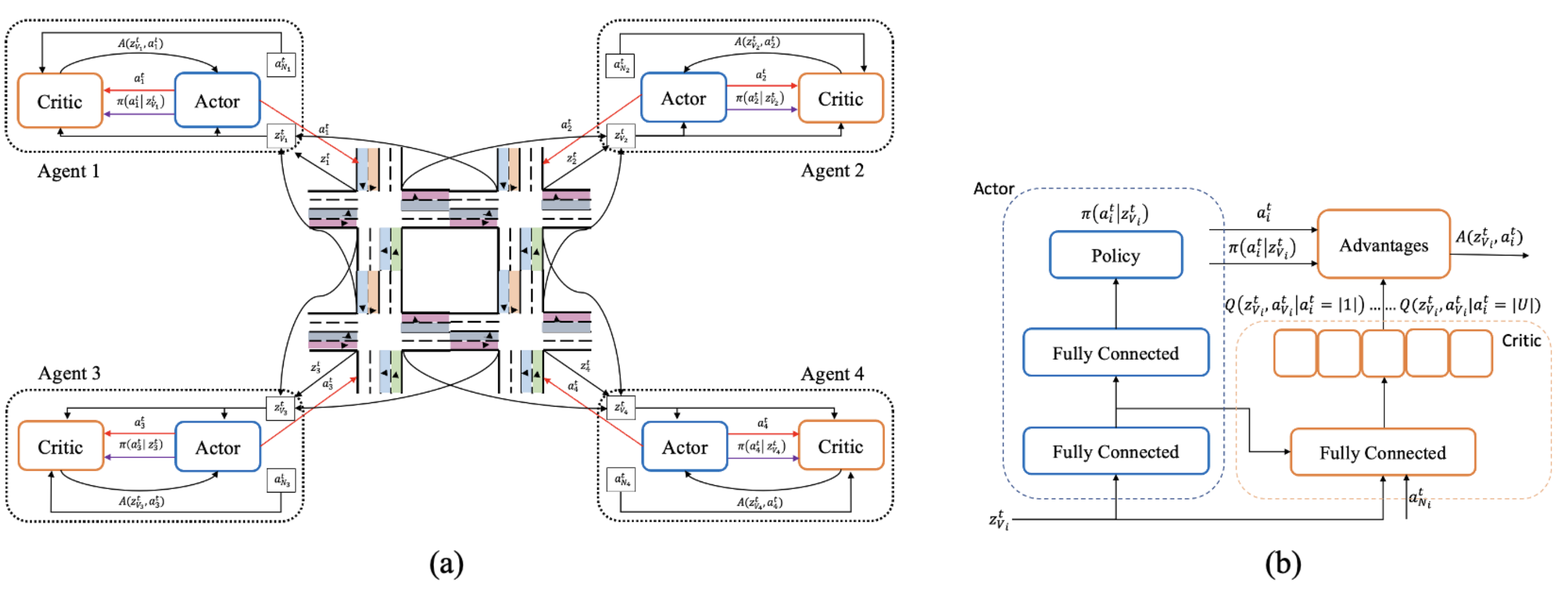}
    \vspace{-0.85cm}
    \caption{Illustration of SocialLight: Fig a) highlights the distributed training framework where each agent maintains an actor network and a local critic network with parameter sharing. The local critic is conditioned on the augmented observations and the actions of neighboring agents to marginalize individual credit via a counterfactual baseline. This baseline is then used to compute the individual advantages used for individual policy improvement. Fig b) Details the actor and critic network architectures. ( Image helped prepared by Stuti Mittal)  }
    \label{fig:my_label}
    \vspace{-0.25cm}
    \Description{Decentralized cooperative learning in SocialLight with locally centralized critics to determine individual contribution marginalization.}
    \label{SocialLightFig}
\end{figure*}

\vspace{-0.25cm}
\subsection{Traffic as a MARL Problem}

\subsubsection{Problem Definition}

Given the current traffic conditions, the goal of traffic agents in the network is to select their own optimal signal phase $a_t$ for a fixed phase duration, until the next decision time-step $t+1$, to maximize a global cumulative objective.

\subsubsection{Observations}

The true global state for a traffic systems comprises the position and current travel time of each vehicle in the network, as well as the current traffic phase of each agent. However, obtaining this information is infeasible in practice. Instead, each agent is only allowed to observe a portion of this full state, which contains the incoming queue lengths, current local traffic phase, average waiting time of queued vehicles, and pressure, which can all be locally measured via Induction Loop Detectors commonly found in modern traffic networks. In this work, aligned with recent work in the community, we use a combination of these features to define each agent's local state.

\subsubsection{Actions}

We let agents directly select one of the 8 traffic phases, and execute that phase for a pre-specified duration (i.e., there is not fixed cycle among phases), for maximal adaptability.
Note other works have also considered switching phases in a fixed cycle without specified phase duration~\cite{marlin-atsc}, or setting phase durations within a fixed cycle length\cite{phase-dur}.

\subsubsection{Rewards}

The global objective for traffic control is to minimize the cumulative trip time of all vehicles.
However, optimizing over total trip time is hard, since vehicles usually accumulate local delays as they pass through multiple junctions in their journey.
Thus, different local reward structures are often used instead, such as cumulative delay, queue length, pressure, or waiting time, which align well enough with the global trip time objective.
In this work, we specifically opt to use queue lengths, to implicitly maximize throughput at each intersection.
This has been recently shown to be superior compared to other local reward formulations~\cite{attentionlight}.
\vspace{-0.25cm}


\section{SocialLight}
\label{method}

SocialLight introduces a new learning mechanism where individual traffic agents learn to cooperate by marginalizing out their true contributions to a neighborhood reward within an independent learning framework. We outline the components proposed in SocialLight to train traffic light control policies for a junction. We first provide an intuition that motivates SocialLight. Then we present the adaptations within the asynchronous actor-critic  framework that are inspired by COMA \cite{coma} to address these challenges. Finally we introduce modifications to the vanilla COMA formulation of the advantages to enhance training stability and improve convergence.

\vspace{-0.2cm}
\subsection{Notation}

Given a traffic network $\mathcal{G(\mathcal{V},\mathcal{E}})$, the local neighborhood for agent $i$ is denoted as $\mathcal{V}_i= i \bigcup \mathcal{N}_i$, where $\mathcal{N}_i$ are the agents in the immediate neighborhood of agent $i$.
Each agent learns a policy network $\pi_{\theta}(a^t_i|z^t_{\mathcal{V}_i})$ and a critic network $Q_{\phi}(z^t_{\mathcal{V}_i},a^t_{\mathcal{V}_i})$, where $z^t_{\mathcal{V}_i}$ is the augmented observation comprising the observation of each agent and its neighbors, i.e., $z^t_{\mathcal{V}_i} = [z^t_{i}] \bigcup_{j \in \mathcal{N}_i} [z^t_{j}]$.
The $a^t_{\mathcal{V}_i}$ denotes the joint action of the agent and it's neighbors i.e. $a^t_{\mathcal{V}_i} = [a^t_{i}] \bigcup_{j \in \mathcal{N}_i} [a^t_{j}]$. 

An agent $i$ receives an individual reward $r^t_{i}$ which can be any reward function computed using local traffic conditions such as queue length over its incoming lanes or local max pressure. Through reward sharing, an individual agent would sum up its own reward with those received by it's neighbors. The neighborhood reward for the agent $i$ is defined as $r^t_{\mathcal{V}_i} = \sum_{j \in \mathcal{V}_i} r^t_{j}$.

\vspace{-0.2cm}
\subsection{Changes to the Policy Gradient}

We leverage the locally centralized critic to compute advantages by marginalizing individual contributions via a counterfactual baseline that is inspired by COMA for policy improvement. COMA learns a central critic $Q(\mathbf{s}^t,\mathbf{a}^t)$ over the global state $s^t$ and global action $\mathbf{a}^t = \bigcup_{i \in N} [a^t_{i}]$ for all $N$ agents. The counterfactual baseline and COMA advantages for a global state is given by $Q(\mathbf{s}^t,\mathbf{a}^t) - \mathop{E}_{a^t_{i}} Q(\mathbf{s}^t,\mathbf{a}^t)$.

In contrast our method computes a local critic $Q_{\phi}(z^t_{\mathcal{V}_i},a^t_{\mathcal{V}_i})$ for each agent i.e. conditioned on the joint observation space and action space of the neighborhood $V_i$ to marginalize individual contribution over the neighborhood reward. Hence in our setting, a naive COMA update is given by
\vspace{-0.1cm}
\begin{equation}
    \hat{A}(z^t_{\mathcal{V}_i},a^t_{i}) = Q_{\phi}(s^t_{\mathcal{V}_i},a^t_{\mathcal{V}_i}) - \sum_{a^t_{i}} \pi_{\theta}(a^t_{i}|s^t_{\mathcal{V}_i}) Q_{\phi}(s^t_{\mathcal{V}_i},a^t_{\mathcal{V}_i}),
\end{equation}

where the second term is a counterfactual baseline that marginalizes an agents expected contributions by fixing the actions of neighboring agents. However, we observed that using original COMA advantages made learning unstable especially with multiple agents updating shared parameters. While the counterfactual baseline's expected contribution to the gradient is zero, we suspect that the instability primarily arises from the high bias in the critic's gradients during the first few epochs of training. To reduce this bias, we introduce two modifications. First, we take inspiration from Temporal Differences (TD), which reduces bias by estimating advantages from rolling out the trajectory and bootstrapping the critic estimates at a future state. However, unbiasing advantages with the estimated return from the full trajectory roll-out comes at the cost of larger variance in the policy gradient. This bias-variance trade-off problem is then further addressed via standard Generalized Advantage Estimation (GAE) over the modified COMA advantages.

TD advantages do not require training an additional network. Hence, we propose a similar modification to the COMA advantages to resembles TD advantages as follows: 
\vspace{-0.1cm}
\begin{equation}
\begin{aligned}
   \hat{A}^1(z^t_{\mathcal{V}_i},a_{t,i}) &= r^t_{\mathcal{V}_i} + \gamma \;\sum_{a^{t+1}_{i}} \pi_{\theta}(a^{t+1}_{i}|z^{t+1}_{\mathcal{V}_i}) Q_{\phi}(z^{t+1}_{\mathcal{V}_i}, a^{t+1}_{\mathcal{V}_i}) \\
   & -  \sum_{a^t_i} \pi_{\theta}(a^t_i|z^t_{\mathcal{V}_i}) Q_{\phi}(z^t_{\mathcal{V}_i},a^t_{\mathcal{V}_i})
\end{aligned}
\end{equation}

Note that our key distinction lies in the estimate of the bootstrapped value which we do by evaluating the counterfactual baseline at the future state. In contrast, TD(1) advantages in a multi-agent setting bootstraps the discounted return over the expectation of the joint action over the policies of all agents $j \in \mathcal{V}_i$ that is given by $\mathop{E}_{\prod_{j \in \mathcal{V}_i} \pi(a^{t+1}_i|z^{t+1}_{\mathcal{V}_i})} Q(z^{t+1}_{\mathcal{V}_i},a^{t+1}_{\mathcal{V}_i})$. However, this fails to capture the dependence of future returns on the future actions of neighboring agents. Hence, setting the bootstrapped return as the counterfactual baseline at the future state reduces the variability of future returns on the actions of future neighboring agents, thereby reducing the variance in the gradient updates to the policy network. 

The above advantages via TD errors still uses a biased critic for the one-step lookahead during the first few epochs. Inspired by the success of GAE, we also use a GAE-type computation to trade off the bias and variance to the policy gradients, for improved learning stability. The generalized advantages are given as:
\vspace{-0.1cm}
\begin{equation}
    A^{GAE}(z^t_{\mathcal{V}_i},a^t_i) = \sum_{l=0}^{\infty} (\gamma \delta)^l \hat{A}^1(z^{t+l}_{\mathcal{V}_i},a^{t+l}_i)
\end{equation}

Note that the Generalised Advantage estimate implicitly weights the $n$ step TD advantages by a factor $\delta^n$ as follows:
\vspace{-0.1cm}
\begin{equation}
\begin{aligned}
   \hat{A}^{n}(z^t_{\mathcal{V}_i},a^t_i) &= \sum_{l=0}^{n-1}\gamma^{l}\; r^{t+l}_{\mathcal{V}_i} + \gamma^{n} \;\sum_{a^{t+n}_i} \pi_{\theta}(a^{t+n}_i|z^{t+n}_{\mathcal{V}_i}) Q_{\phi}(z^{t+n}_{\mathcal{V}_i}, a^{t+n}_{\mathcal{V}_i}) \\
   & -  \sum_{a^{t}_i} \pi_{\theta}(a^t_i|z^t_{\mathcal{V}_i}) Q_{\phi}(z^t_{\mathcal{V}_i},a^t_{\mathcal{V}_i})
\end{aligned}
\end{equation}

The discounting factor $\delta$ regulates the bias variance tradeoff where the variance in the gradient estimator increases with the time horizon due to the influence of the returns on the actions of the team in the neighborhood. The policy gradient for a rollout of length $T$ is thereby given as:

\begin{equation}
    \nabla L_{\pi}(\theta) = \sum_{t=0}^{T} \nabla_{\theta}\; log(\pi_{\theta}(a_t^i|z^t_{\mathcal{V}_i}) \; A^{GAE}(z^t_{\mathcal{V}_i},a^t_i)
\end{equation}

\subsection{Critic Training}

The critic introduced in the above section estimates returns over a joint action space of the agents in the neighborhood $\mathcal{N}_i$. However, having the critic output $\mathcal{|A|}^n$ values, where $\mathcal{|A|}$ represents the size of the action space of one agent, is impractical. We address this problem by using a critic representation similar to COMA which allows for an efficient evaluation of the baseline.
In this work, the critic is a neural network that takes local observations of the neighborhood $z^t_{\mathcal{V}_i}$ and the actions of other agents $a^t_{\{\mathcal{V}_i\; - i\}}$  and outputs the a vector of length $|A|$.
Note that the actions of other agents are one hot encoded; the network is depicted in Fig. ~\ref{SocialLightFig}.
In doing so, the advantage term can be computed in a single pass through a dot product between the outputs of the actor and critic networks.

To conform with the TD
error used to compute advantages via the counterfactual baseline computed at the joint future agent state, the targets of the critic are modified similarly.
Here the TD(1) error is given by
\begin{equation}
    G^t_i = r^t_{\mathcal{V}_i} + \gamma \;\sum_{a^{t+1}_i} \pi_{\theta}(a^{t+1}_i|z^{t+1}_{\mathcal{V}_i}) Q_{\phi}(z^{t+1}_{\mathcal{V}_i}, a^{t+1}_{\mathcal{V}_i})
\end{equation}

More generally, the $n$ step TD returns is formulated as
\begin{equation}
    G^{t:t+n}_i = \sum_{l=0}^{n-1}\gamma^{l}\; r^{t+l}_{\mathcal{V}_i} + \gamma^{n} \;\sum_{a^{t+n}_i} \pi_{\theta}(a^{t+n}_i|z^{t+n}_{\mathcal{V}_i}) Q_{\phi}(z^{t+n}_{\mathcal{V}_i}, a^{t+n}_{\mathcal{V}_i})
\end{equation}

The TD($\lambda$) targets $G_{t,i}^{\lambda}$ can be computed via these modified $n$ step returns and the critic network is regressed against these targets over a rollout of length $T$. The critic loss reads
\begin{equation}
    L_Q(\phi) = \frac{1}{T} \sum_{t=0}^T (Q_{\phi}(z_{t+n,\mathcal{V}_i}, a_{t+n,\mathcal{V}_i}) - G_{t,i}^{\lambda})^2
\end{equation}



\section{Experiments}
\label{experiments}

\begin{figure*}[htp]
    \centering
    \vspace{-0.4cm}
	\subfloat{ \includegraphics[height=1.5in]{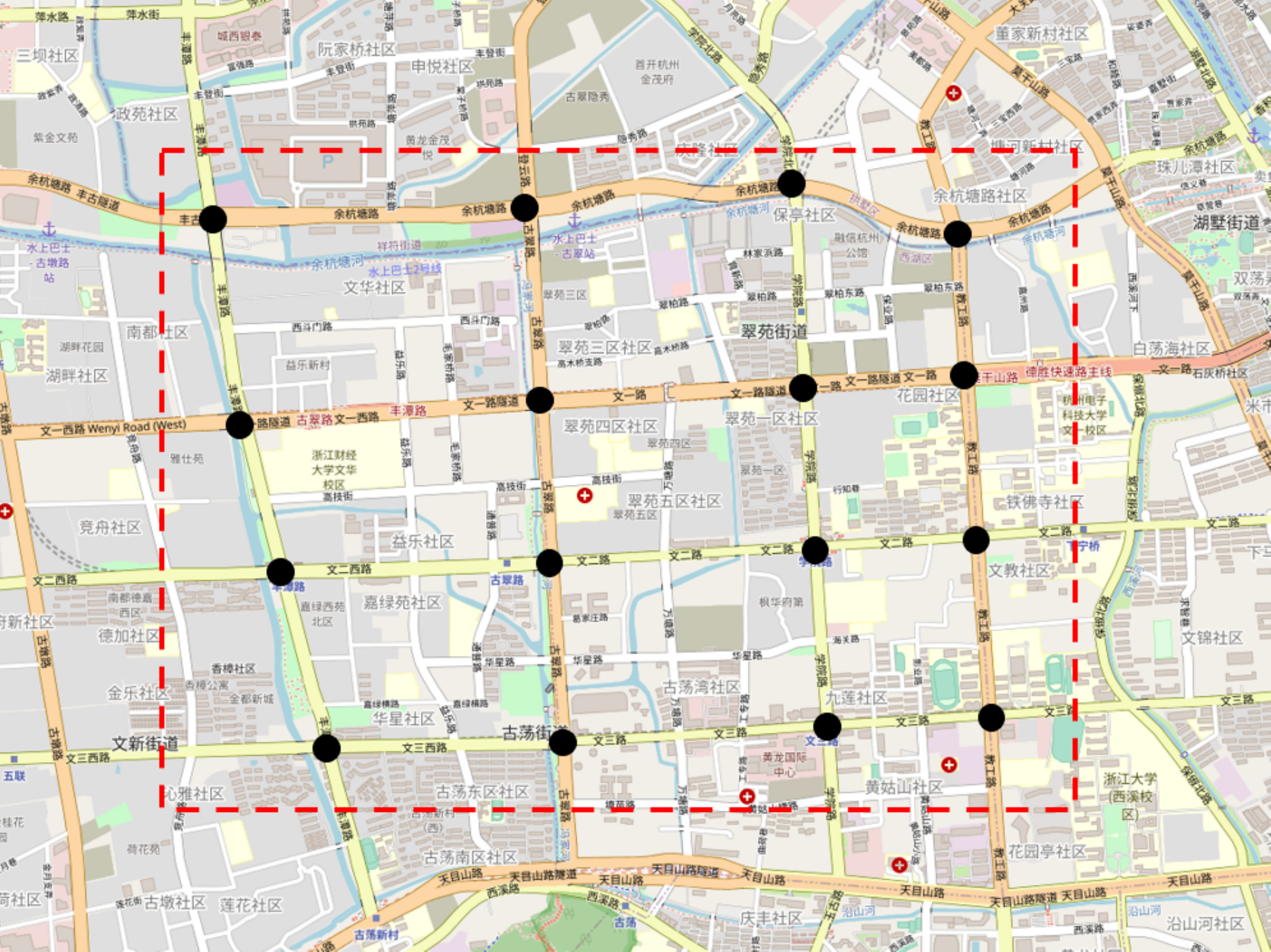}}
	\hspace{1em}
	\subfloat{ \includegraphics[height=1.5in]{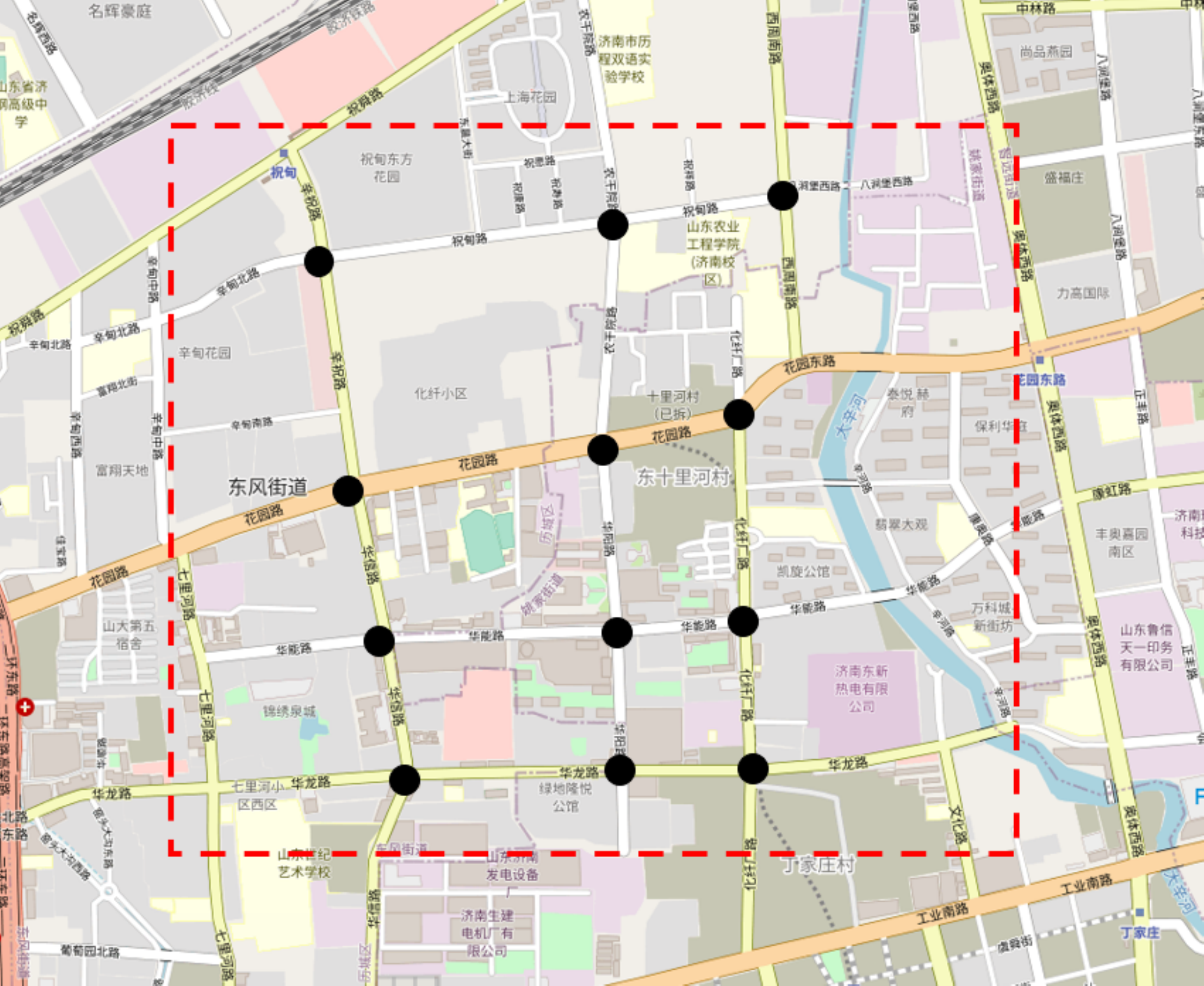}}
	\hspace{1em}
	\subfloat{ \includegraphics[height=1.5in]{New_York.pdf}}
	\hspace{1em}
    \caption{Real-world traffic road network for CityFlow dataset (From left to right, Hangzhou map, Jinan map and New York map)}
    \Description{Real-World traffic road networks for CityFlow datasets (From left to right, Hangzhou map, Jinan map and New York map)}
\label{network_visualisation}
\vspace{-0.2cm}
\end{figure*}

\begin{table*}[t]
\renewcommand{\arraystretch}{1.3}
\caption{SocialLight vs Baselines on CityFlow}
\centering

\begin{tabular}{l||c|c|c|c|c|c|c}
\midrule
{\bf Methods} & \multicolumn{2}{c}{\bf New York}  & \multicolumn{2}{|c|}{\bf Hangzhou} & \multicolumn{3}{c}{\bf Jinan} \\
\cmidrule(lr){2-8}
 & {\bf 1} & {\bf 2} & {\bf 1} & {\bf 2 }& {\bf 1} & {\bf 2} & {\bf 3}\\
\midrule
Fixed Time & 1397.37 & 1660.29 & 432.316 & 359.44 & 364.36 & 289.74 & 316.69\\
Max Pressure &  1177.75 & 1535.77 & 262.35 & 348.68 & 275.79 & 223.06 & 234.93 \\
Co-Light &  1221.77 & 1476.18 & 271.07 & 297.26 & 276.33 & 237.14 & 278.16 \\

MP - Light &  1168.49 & 1597.24 & 343.47 & {\bf 282.14} & 300.93 & 259.10 & 261.45\\
Attention-Light &  978.62 & 1571.68 & 259.62 & 284.75 & 268.68 & 212.76 & 216.34
 \\
SocialLight  & {\bf 760.94} & {\bf 1114.47} & {\bf 255.68} & 301.16 & {\bf 226.21} & {\bf 209.81} & {\bf 205.38}
 \\
\midrule
\midrule
Performance Gain & 22.23 \% & 24.53 \% & 1.54 \% & - & 15.68\% & 1.41 \% & 5.09 \%\\
\end{tabular}
\label{table-cf}

\end{table*}

Our experiments aim to answer the following questions:

\begin{enumerate}
    \item Does SocialLight improve over the current state of the art, when keeping standard definitions of the state space and reward?
    \item What is the impact of the proposed individual contribution marginalization mechanism in comparison to standard reward sharing?
    \item Do our modified advantages improve the stability of training, compared to simply applying COMA advantages locally?
\end{enumerate}
    
To answer these questions, our experiments are conducted in both the SUMO and CityFlow traffic simulators~\cite{tang2019cityflow,krajzewicz2010traffic}.
We first benchmark the performance of SocialLight on common baselines developed over both the synthetic SUMO and real world Cityflow traffic datasets and measure the same standard traffic metrics in both datasets. 

We then perform an ablation to study the impact of individual contribution marginalization on the traffic performance over  artificially generated traffic flows on a Manhattan road-map over the SUMO simulator. By simulating a diverse range of traffic scenarios to train and test on, we can eliminate networks potentially over-fitting to a single scenario that may misrepresent our analysis.
Moreover, we show the improvement over the learning stability of our network over both synthetic traffic and a real traffic datasets. 

\vspace{-0.2cm}
\subsection{Description of Traffic Datasets}

We conduct experiments on two different microscopic traffic simulators SUMO~\cite{SUMO2018} and CityFlow~\cite{tang2019cityflow} with synthetic and real-world datasets respectively.
A traffic simulation on either simulator comprises a road network and a traffic flow dataset. Here the road network defines the positions of the intersections, the attributes of the roads (e.g., number and length of lanes, speed limits, and lane connections etc.) and the phase settings.
The traffic flow datasets define the travel information of all vehicles, characterized by the origin destination (O-D) pair and time that the vehicle enters the network.
Note that our simulations are conducted over homogeneous intersections that have the same settings for the roads and traffic light phases.

\vspace{-0.20 cm}
\subsubsection{Synthetic Traffic Datasets}

We use synthetic traffic dataset with a Manhattan road network on the SUMO simulator that is adapted from the benchmark method MA2C~\cite{ma2c}. The road network is a  $5 \times 5$ traffic grid network with 25 intersections. Each intersection is formed by two-lane streets (W-E) with speed limit 72 km/h and one-lane avenues with speed limit 40km/h. 

The traffic flow datasets are artificially generated during run time with different fixed seeds, to fairly compare among algorithms.
Aligned with previous work~\cite{ma2c}, all episodes consider a fixed traffic flow composed of travel information (O-D pair) for all vehicles, while the seed sets the (random) initial position and speed of these vehicles.

\vspace{-0.20 cm}
\subsubsection{Real Traffic Datasets}

The real traffic dataset considers three city networks - Jinan, Hangzhou and New York. These datasets serve as popular benchmarks for ATSC~\cite{zheng2019frap,wei2019colight, wei2019survey}. 
The road networks are extracted from a portion of real-world traffic map for simulations, and traffic flow datasets were compiled from cameras during different time periods.
As shown in Figure~\ref{network_visualisation}, there are a total of 12 (3$x$4) intersections in the Jinan map, 16 (4$x$4) intersections in the Hangzhou map, and 192(28$x$7) intersections in the New York map.
The traffic flow datasets used in these different maps are described in~\cite{attentionlight}; there are three different flow datasets for Jinan, two for Hangzhou, and two for New York.


\begin{table*}[t]
\renewcommand{\arraystretch}{1.5}
\small
\caption{SocialLight vs Baselines on SUMO - Manhattan }
\label{table_example}
\centering

\begin{tabular}{l||c|c|c|c|c|c|c}
\midrule
{\bf Metrics(Average)} & {\bf Greedy} & {\bf IA2C} & {\bf MA2C} & {\bf IQL-LR}  & {\bf A3C }& {\bf A3C (nr)} & {\bf SocialLight}\\
\midrule
Queue Length & 5.00 (2.86) &
3.27 (2.00) & 2.24 (1.28)& 3.76(2.74)  & 1.71 (1.14) & 1.11 (0.89) &{\bf 0.74 (0.69)}\\
Speed (m/s)  & 1.56 (1.33)  & 1.70 (1.32) & 2.31 (1.22)& 3.04(3.24)  & 4.55 (2.78) & 4.94 (2.34) & {\bf 5.36 (2.66)} \\
Intersection Delay (s)& 60.97 (47.15) & 58.27 (46.08)  & 21.96 (19.83)& 92.59(94.69)  & 38.23 (32.83) & 25.43 (22.53) & {\bf 10.07 (9.02)} \\
Cumulative Delay (s) & 595.78 (464.23) & 446.85 (410.71) & 325.02 (269.77) & 214.521(342.95) & 184.60 (297.99) & 159.23 (226.64) & {\bf 106.51 (165.18)}\\
Trip Time (s) & 885.47 (572.63) & 704.62 (501.33) & 597.88 (399.09) & 462.862(453.69) & 395.04 (387.06) & 386.01 (305.74) & {\bf 309.81 (243.95)} \\
\midrule
\end{tabular}
\label{table-manh}
\end{table*}

\vspace{-0.20cm}
\subsection{Traffic Performance on SUMO Synthetic Traffic Datasets}

We compare our method SocialLight with current state-of-the-art baselines optimized on the various traffic datasets. 
Following the baselines \cite{ma2c} for a fair comparison, we keep the same experiment settings, as well as similar POMDP settings in terms of the definitions of actions and rewards. 
The state is modified to include current traffic phase.

\subsubsection{Baseline methods}
\begin{enumerate}
    \item \textbf{Greedy~\cite{koonce2008traffic}}:
    Greedily chooses the phase associated with lanes with maximum incoming queue length.
    
    \item \textbf{IQL-LR~\cite{ma2c}}: A linear regression based independent Q-learning (IQL) algorithm, where each local agent learns its own policy independently by considering other agents as part of the environment's dynamics.
    
    \item \textbf{IA2C~\cite{ma2c}}: An extension of IQL-LR which relies on advantage actor-critic (A2C) algorithm instead of IQL.

    \item \textbf{MA2C~\cite{ma2c}}: A cooperative MARL algorithm, which includes the observations and fingerprints of neighboring agents in each agent's state.

    to alleviate the instability caused by partial observability. MA2C further introduces a spatial discount factor to scale down the observation and rewards signals of the neighboring agents, to encourage agents towards neighborhood-level cooperation.
    
    \item \textbf{A3C}: The distributed learning framework with parameter sharing relying on A3C algorithm, where each agent learns to maximize its individual objective.
    
    \item \textbf{A3C(nr)}: The distributed learning framework where each agent is to maximize the neighborhood reward rather than individual reward.
    
\end{enumerate}

\begin{figure*}[htp]
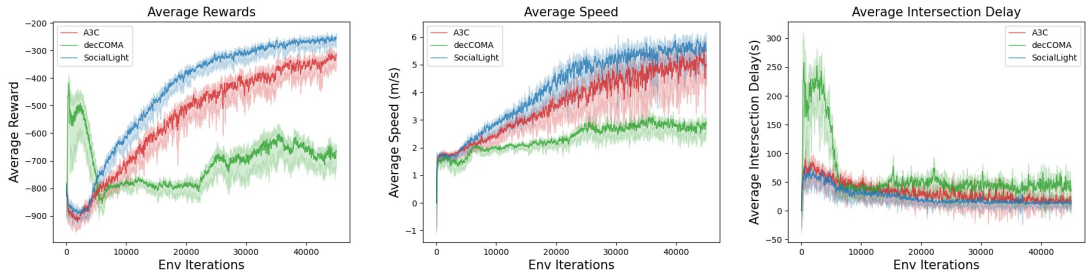

    \vspace{-0.4cm}
    \centering
    \subfloat{ \includegraphics[height=1.5in]{plots_rewards_Manh.pdf}}
	\hspace{-1em}
	\subfloat{ \includegraphics[height=1.5in]{plots_speed_Manh.pdf}}
	\hspace{-1em}
	\subfloat{\includegraphics[height=1.5in]{plots_delay_Manh.pdf}}
	\hspace{-1em}
    \caption{Training Plots of methods over average rewards, vehicle speed and intersection delay on the SUMO Manhattan synthetic data-set. SocialLight is shown in Blue, A3C with Neighborhood Rewards is in red and SocialLight with original COMA advantages which we refer to as decCOMA is in green. Observe that decCOMA fails to improve over cumulative returns. }
    \Description{Training Plots of methods over average rewards, vehicle spped and intersection delay on the SUMO Manhattan synthetic dataset.}
\label{simresults}
\end{figure*}  
\vspace{-0.2cm}
\subsubsection{Analysis}

We first observe that SocialLight outperforms the heuristic based Greedy baseline, as well as the Deep RL baselines MA2C, IA2C, IQL-LR and even A3C with and without neighborhood reward over all traffic metrics (average queue length, speed, intersection delay, cumulative delay, and average total trip time).
Prior methods IA2C, MA2C and IQL-LR are outperformed significantly by both  A3C with and without neighborhood rewards, even though these are on policy actor-critic methods with the same POMDP settings.
We believe that this may be due to the way in which neighboring states are aggregated into each agents' individual state via \textit{discounted} summation in these baselines, which results in poorer policies.
SocialLight, on the other hand, significantly outperforms standard A3C methods with and without neighborhood reward in terms of average trip time and cumulative delays. This is most likely due to the proposed contribution marginalization scheme within the neighborhood reward which tightly couples the given agent with neighboring agents to maximize throughput across the neighborhood, thereby reducing travel time and delays.

\vspace{-0.20cm}
\subsection{Traffic Performance on CityFlow Real Traffic Datasets}
We evaluate SocialLight on real traffic datasets. The experiment settings and POMDP settings are unchanged with respect to \cite{attentionlight} for a fair comparison among all methods.
\subsubsection{Baseline methods}

\begin{enumerate}
    \item \textbf{FixedTime~\cite{koonce2008traffic}}:
    Fixed time control considers a fixed cycle over phases with a pre-defined total phase length and pre-defined phase split over the total cycle length.
    \item \textbf{MaxPressure~\cite{varaiya2013max}} :
    MP (max-pressure) control greedily selects the phase that can minimize the intersection pressure, where the pressure is calculated by the difference between the vehicles of incoming lanes and connected outgoing lanes.
    \item \textbf{Co-Light~\cite{wei2019colight}} :
    A state-of-the-art method that uses Graph Attention Neural Networks to accomplish junction level cooperation and has been trained via Deep Q learning.
    \item \textbf{MP-Light~\cite{chen2020toward}}
    MP-Light incorporates pressure in their states and reward to achieve state-of-the art scalable ATSC over a city-level traffic network. MP-Light also applies the FRAP based training architecture that uses phase competition within various traffic movements to improve control performance.
    \item \textbf{Attention-Light}
    Attention-Light~\cite{attentionlight} is a recent state-of-the-art model that incorporates self-attention to learn the phase correlation and competition in contrast to FRAP that uses human knowledge. This method has shown to outperform FRAP based MPLight and the CoLight over numerous traffic datasets.
    
\end{enumerate}
\vspace{-0.20 cm}
\subsubsection{Analysis}

Our results show that SocialLight in Table \ref{table-cf} outperforms all existing baselines over a wide variety of traffic flows in terms of reducing average travel time.
In particular, we observe the highest performance gains over the New York traffic flow sets that comprises 196 traffic agents.
We report more modest performance gains over the Hangzhou and Jinan traffic flow datasets.

Over traffic flow datasets compiled for Jinan and Hangzhou urban networks, SocialLight shows modest improvements compared to the state-of-the-art AttentionLight.
Despite large gains on NewYork, the relatively modest performance gains of SocialLight over the Hangzhou and Jinan can be attributed to the saturation. This saturation results from their relatively small scale in terms of the number of agents in the training datasets. This saturation is evident from the performance classical baselines such as MaxPressure over these traffic flow. 
For instance, the average trip time performance is actually observed to degrade with Co-Light and MP-Light as compared to the MaxPressure baseline on the Hangzhou and Jinan datasets. Attention-light is found to have very marginal performance gains as compared to the MaxPressure baseline (except for Hangzhou-2).
This observations indicate that any potential performance gains from these datasets would be minimal due to saturation. Hence, SocialLight performs better than AttentionLight and the MaxPressure baselines on most traffic datasets. 

We then turn our attention to the New York traffic flow sets, which are harder to learn over due to their sheer scale.
Of the two flow sets there, New York-2 is the hardest traffic dataset~\cite{attentionlight} in terms of vehicle arrival rate.
Most RL methods (CoLight, MPLight and AttentionLight) produce modest performance gains compared to the classical FixedTime and MaxPressure baselines on both the New york traffic flowsets.
Co-light improves over the New York-2 dataset compared to the classical methods, but fails to do so over the New York-1 dataset.
In contrast, AttentionLight fails to improve over the classic MaxPressure controller in the NewYork-2 dataset.
We believe that these improvements indicate the need for true cooperation, which Co-Light and AttentionLight only achieve via their network designs. 

SocialLight pushes the state-of-the-art in terms of traffic performance on both these New York datasets, where our performance gains over the current state-of-the-art AttentionLight and CoLight are even more pronounced ($\geq 20 \%$).
We believe that this is due to the impact of agents marginalizing individual contributions to the neighborhood reward, which encourages better cooperation as each agent understands its role in the local traffic performance.
By overlapping the agents' neighborhoods (i.e., local areas of enhanced cooperation), our approach effectively leads to improved network-wide cooperation and thus large performance gains.

\vspace{-0.25cm}
\subsection{Impact of Individual Contribution Marginalization on Learning and Final Performance}

This ablation study aims to identify the impact of contribution marginalization on both the learning process and the final performance of trained policies.
We compare SocialLight with vanilla Asynchronous Actor Critic (A3C) with reward sharing (rewards of neighbors summed up).
As shown in Fig.~\ref{simresults}, SocialLight exhibits improved sample efficiency and improved returns as compared to A3C even though both algorithms are on-policy and have the same architectures and augmented state inputs for their policy networks. 

Furthermore, we note that decentralized contribution marginalization introduced in SocialLight is also shown to have improved the stability of training.
While both algorithms converge, Fig~\ref{simresults} indicates larger variances in the returns, average speed and intersection delays during training for A3C with neighborhood reward sharing. 

We further compare both the networks over a validation set of synthetic traffic flows generated from separate seeds that generate traffic flows during training.
The improvement over the total trip time and cumulative delay shown in Table~\ref{table-manh} highlight how the marginalization of individual contributions promote cooperation between traffic agents for large performance gains.

\vspace{-0.20cm}
\subsection{Impact of Modified Advantages on Training Stability and Convergence}

Further analysis shows the impact of the modified advantages that marginalizes individual contributions.
Compared to simply applying COMA advantages~\cite{coma}, our modifications are shown to improve training stability as shown in Fig~\ref{simresults} and overall returns.
We believe that original COMA advantages converge to sup-optimal policies due to the highly distributed nature of our training process, where high initial biases in the critic inhibit monotonic policy improvement during the initial phase of training.
As both the policy and networks learn in conjunction, the policy network learned via the original COMA advantages converge to sub-optimal traffic control policies, differently from SocialLight.

\vspace{-0.1cm}
\section{Conclusion}
\label{conclusion}
This paper presents SocialLight, a fully decentralized training framework that learns cooperative traffic light control policies via distributedly marginalizing individual contributions to each agent's local neighborhood reward.
We show that our method improves over the scalability for cooperative learning, thereby improving final traffic performance (average trip time) especially over large traffic networks such as the New York grid with 196 traffic intersections.
These performance gains suggest that our method could improve the overall quality of learned policies on real-life citywide networks.

Our method leverages the fixed spatial structure of traffic systems to define overlapping neighborhoods over which agents can marginalize their contributions, to scale cooperative learning without the need for a centralized critic.
Future work will focus on extending this idea to general mixed competitive-cooperative games, where such distributed spatial structures are more difficult to identify and leverage.
There, we will aim to develop methods to learn suitable neighborhoods over which individual agents can marginalize their contributions over to improve the scalability of cooperative learning.



\newpage
\begin{acks}

This work was partly supported by A*STAR, CISCO Systems (USA) Pte. Ltd and National University of Singapore under its Cisco-NUS Accelerated Digital Economy Corporate Laboratory (Award I21001E0002).

\end{acks}




\bibliographystyle{ACM-Reference-Format} 
\balance
\bibliography{sample}
\vfill\eject
\newpage
\section{Supplemental Material}
\subsection{Experiment Settings}
\subsubsection{SUMO Dataset}
Each simulation lasts for 3600 seconds, where the phase duration $\Delta T$ is fixed to 5 seconds. Hence there are 720 RL time steps in a single simulation episode. After a traffic phase change, the yellow time $\Delta t_y$ is set to 2s~\cite{ma2c}.

\subsubsection{CityFlow dataset}
We adapt the same traffic phase duration and yellow duration settings as SUMO for the simulations.

\subsection{MDP Settings}

\subsubsection{SUMO Dataset}
The Markov Decision Process Settings(State, Actions and Rewards) are defined as follows:

\begin{enumerate}
\item \textbf{Actions}:
The traffic signal directly controls one of the 5 available phases of the traffic light~\cite{ma2c}.
Each agent executes a phase for a fixed duration in the simulator as specified by the total phase duration $\Delta T$.

\item \textbf{Observation}:
    
The local state of an individual agent is defined by its current traffic phase, waiting time, and traffic queue length at each incoming lane and outgoing lane at the traffic intersection.
Here, the waiting time is the normalized cumulative delay of the first vehicle along an incoming lane.
The outgoing traffic queue lengths and incoming traffic queue lengths are obtained from near-intersection induction-loop
detectors (ILD). 

\item \textbf{Rewards}:
Agent rewards are set to the incoming queue lengths over the lane area  detectors in SUMO. These detectors measure incoming traffic upto a certain distance from the intersection. This has been a standard reward structue for many prior works \cite{attentionlight,ma2c,chu2019multi}.

\end{enumerate}

\subsubsection{CityFlow Dataset}
The Markov Decision Process Settings(State, Actions and Rewards) are defined as follows:

\begin{enumerate}

\item \textbf{State}:
Following~\cite{attentionlight}, the local state of the agent comprises the one-hot encoded current phase (action) taken by the agent and the queue length across each incoming traffic lane at an intersection.   
\item \textbf{Action}:
The traffic agent executes one of the 8 phases as described in \ref{intersection}.

\item \textbf{Rewards}:
Following~\cite{attentionlight}, agent rewards are set to the negative of the queue lengths over incoming traffic lanes in the intersection. Optimizing this reward metric maximizes the throughput from the given intersection. With reward sharing, each traffic agent would then intend to maximize the throughput through the local neighborhood. 
\end{enumerate}

\end{document}